\documentclass[conference, 10pt]{IEEEtran}

\usepackage{amsmath, amsthm, amssymb, amsfonts}

\usepackage{algorithmic}
\usepackage{algorithm}

\usepackage[pdftex]{graphicx}
\ifCLASSOPTIONcompsoc
  \usepackage[caption=false,font=normalsize,labelfont=sf,textfont=sf]{subfig}
\else
  \usepackage[caption=false,font=footnotesize]{subfig}
\fi
\usepackage{array,multirow, booktabs}
\usepackage{amsmath}

\newcolumntype{C}[1]{>{\centering}p{#1}}

\usepackage{tikz}
\usetikzlibrary{positioning}
\usetikzlibrary{calc}
\usetikzlibrary{patterns}

\tikzset{%
	dblock/.style = {rectangle, 
		text centered, dashed, draw,
		semithick, rounded corners,
		font=\scriptsize,
	},
	block/.style = {rectangle, draw, 
		font=\scriptsize,
		text width=6.5em, text centered,
		rounded corners, minimum height=2em 
	},
	tblock/.style = {rectangle,
		font=\scriptsize,
	},
	diagonal fill/.style 2 args={draw, rounded corners,
		font=\sffamily\footnotesize,
		text width=8em, text centered, fill=#2, path picture={
			\fill[#1, sharp corners] (path picture bounding box.south west) -|
			(path picture bounding box.north east) -- cycle;}},
}

\usepackage{pgfplots}
\usepgfplotslibrary{fillbetween}
\usetikzlibrary{pgfplots.groupplots}
\pgfplotsset{compat=newest, 
	tick label style={font=\scriptsize},
	label style={font=\scriptsize},
	legend style={font=\scriptsize, row sep=-1pt},
every non boxed x axis/.style={} 
}

\newenvironment{customlegend}[1][]{%
	\begingroup
	\csname pgfplots@init@cleared@structures\endcsname
	\pgfplotsset{#1}%
}{%
	\csname pgfplots@createlegend\endcsname
	\endgroup
}%

\def\addlegendimage{\csname pgfplots@addlegendimage\endcsname}

\pgfplotscreateplotcyclelist{custom}{%
	blue!50!black,every mark/.append style={fill=blue!80!black},mark=*\\
	red!70!black,densely dashed,every mark/.append style={solid, fill=red!80!black},mark=star,mark options={scale=2}\\
	green!50!black,densely dotted, mark=square*,every mark/.append style={solid,fill=green!60!black}\\
   darkgray,dashdotted,every mark/.append style={solid,fill=darkgray},mark=diamond*\\
}

\pgfplotscreateplotcyclelist{histcustom}{%
	blue!50!black,fill=blue!90!white\\
	red!70!black,fill=red!20!white,postaction={pattern=horizontal lines, pattern color=red!70!black}\\
	green!50!black,fill=green!60!white,postaction={pattern=north east lines, pattern color=green!50!black}\\
	yellow!50!black,fill=yellow!50!white\\
	darkgray,fill=lightgray,postaction={pattern=dots, pattern color=darkgray}\\
	cyan!60!black,fill=cyan!60!white\\
}

\begin{document}

\onecolumn
%
%
%

\vspace*{3cm}

{\Large IEEE Copyright Notice}

\vspace*{0.5cm}

Copyright $\copyright$ 2020 IEEE
Personal use of this material is permitted. Permission from IEEE must be obtained for all other uses, in any current or future media, including reprinting/republishing this material for advertising or promotional purposes, creating new collective works, for resale or redistribution to servers or lists, or reuse of any copyrighted component of this work in other works.

\vspace*{1cm}

{\Large\textbf{A Privacy-Preserving  Distributed Architecture for Deep-Learning-as-a-Service}}

\vspace*{0.5cm}

Simone Disabato\footnotemark[1]

Alessandro Falcetta\footnotemark[1]

Alessio Mongelluzzo\footnotemark[1]

Manuel Roveri\footnotemark[1]

\footnotetext[1]{Politecnico di Milano, Dipartimento di Elettronica, Informazione e Bioingegneria, Milano, Italy.}

\vspace*{0.5cm}

{\large Accepted to be published in:  2020 International Joint Conference on Neural Networks (IJCNN), Glasgow, July 19--24, 2020}


\vfill

{\Large Please cite as:}

\vspace*{0.5cm}

\begin{minipage}[l]{0.75\textwidth}
    S. Disabato, A. Falcetta, A. Mongelluzzo, and M. Roveri. "A Privacy-Preserving  Distributed Architecture for Deep-Learning-as-a-Service." 2020 International Joint Conference on Neural Networks (IJCNN). IEEE, 2020.
    
\end{minipage}
    
\vfill

{\Large BibTex}

\vspace*{0.5cm}

%
%
%
%
%
%
%
%

\begin{minipage}[l]{0.75\textwidth}
    
    \begin{verbatim}
@inproceedings{disabato2020privacy,
    title={A Privacy-Preserving  Distributed Architecture for
           Deep-Learning-as-a-Service},
    author={Disabato, Simone, and Falcetta, Alessandro, and
            Mongelluzzo, Alessio, and Roveri, Manuel},
    booktitle={2020 International Joint Conference on Neural
               Networks (IJCNN)},
    pages={1--8},
    year={2020},
    organization={IEEE}
}
    \end{verbatim}

\end{minipage}

\vfill

\clearpage

\twocolumn

\bstctlcite{IEEEexample:BSTcontrol}
%
%
\title{\huge A Privacy-Preserving  Distributed Architecture\\for Deep-Learning-as-a-Service}

\author{\IEEEauthorblockN{Simone Disabato, Alessandro Falcetta, Alessio Mongelluzzo, and Manuel Roveri}
\IEEEauthorblockA{Dipartimento di Elettronica, Informazione e Bioingegneria,\\
Politecnico di Milano, Milano, Italy\\
 Email: alessandro.falcetta@mail.polimi.it, \{simone.disabato,alessio.mongelluzzo,manuel.roveri\}@polimi.it}
}


%


\maketitle

\begin{abstract}
Deep-learning-as-a-service is a novel and promising computing paradigm aiming at providing machine/deep learning solutions and mechanisms through Cloud-based computing infrastructures. Thanks to its ability
to remotely execute and train deep learning models (that typically require high computational loads and memory occupation), such an approach guarantees high performance, scalability, and availability. Unfortunately, such an approach requires to send information to be processed (e.g., signals, images, positions, sounds, videos) to the Cloud, hence having potentially catastrophic-impacts on the privacy of users. This paper introduces a novel distributed architecture for deep-learning-as-a-service that is able to preserve the user sensitive data while providing Cloud-based machine and deep learning services. The proposed architecture, which relies on Homomorphic Encryption that is able to perform operations on encrypted data, has been tailored for Convolutional Neural Networks (CNNs) in the domain of image analysis and implemented through a client-server REST-based approach.  Experimental results show the effectiveness of the proposed architecture.
\end{abstract}

%
\IEEEpeerreviewmaketitle

\section{Introduction}
\label{sct:introduction}
In recent years, the technological evolution of Cloud-based computing infrastructures intercepted the ever-growing demand of machine and deep-learning solutions leading to the novel paradigms of \textit{machine} and \textit{deep-learning-as-a-service}~\cite{yao2017complexity}. The core of such computing paradigms is that Cloud providers provide ready-to-use remotely-executable machine/deep learning services in addition to virtual computing environments  (as in infrastructure-as-a-service) or platform-based solutions (as in platforms-as-a-service). Examples of such services are the identification of faces in images or videos or the conversion of text-to-speech or speech-to-text~\cite{hossain2015cloud}. From the perspective of the user, being ready-to-use, these services do not require the training of the models (that are pre-trained by the Cloud provider) nor the local recall of such models (that are executed on the Cloud). Moreover, the Cloud-based computing infrastructure providing such machine/deep learning solutions  \textit{as-a-service} allows to support scalability, availability, maintainability, and pay-per-use billing mechanisms~\cite{erl2013cloud}. 

Unfortunately, to be effective, such an approach involves the processing of data that might be sensitive, e.g., personal pictures or videos, medical diagnoses, as well as data that might reveal ethnic origin, political opinions, but also genetic, biometric and health data~\cite{gdpr2016personal}.

The aim of this paper is to introduce a novel distributed architecture meant to preserve the privacy of user data in the deep-learning-as-a-service computing scenario. To achieve this goal, the proposed architecture relies on Homomorphic Encryption (HE) that is an encryption scheme allowing the process of encrypted data~\cite{acar2018survey}. In the proposed architecture, by exploiting the properties of HE, users can locally encrypt their data through a public key, send them to a suitably-encoded Cloud-based deep-learning service (provided through the deep-learning-as-a-service approach), and receive back the encrypted results of the computation that are locally decrypted through the private key. More specifically, such architecture allows to decouple the encryption/decryption phases, which are carried out on the device of the user (e.g., a personal computer or a mobile device), from the deep-learning processing, which is carried out on the Cloud-based computing infrastructure. Such a HE-based distributed architecture allows to preserve the privacy of data (plain data are never sent to the Cloud provider) while guaranteeing scalability, availability, and high performance provided by Cloud-based solution. 

The ability to process encrypted data of HE comes at two main drawbacks. First, the computational load and the memory demand of HE-encoded operations is much higher than regular ones, hence making the HE-encoded deep-learning processing highly demanding in terms of computation and memory. This is the reason why we focused on a deep-learning-as-a-service approach where the computation is carried out on high performing units on the Cloud. Second, HE supports only a limited set of operations (typically sums and multiplications). For this reason, prior to the encoding provided by the HE scheme, the deep-learning models have to be redesigned and retrained taking into account the constraints on the set of available operations. In addition, HE schemes have to be configured through some parameters that trade-off the accuracy in the computation with the computational loads and memory occupation. Such a configuration, that depends on the processing chain and the data to be processed, is managed at the Cloud-level by providing different settings of parameters that can be explored by the user.

The proposed architecture is intended to work with any machine and/or deep learning solution. However, in this work, it has been tailored to image analysis solutions leveraging Convolutional Neural Networks (CNNs)~\cite{lecun1995convolutional}, and implemented through a client (locally executed on the user device) developed as a Python library and a server developed as a deep-learning-as-a-service container implemented on Amazon AWS. The developed architecture relies on a Representational state transfer (REST) paradigm for exchanging encrypted data and results between client and server, while messages rely on JSON format.

A wide experimental campaign shows the feasibility and evaluates the performance of the proposed architecture. The Python Library for the client and the Amazon AWS Container are made available to the scientific community\footnote{Code is available for download as a public repository at https://github.com/AlexMV12/PyCrCNN.git}.

The paper is organized as follows. Section~\ref{sct:background} introduces a background on HE, while 
Section~\ref{sct:literature} describes the related literature. The proposed architecture is detailed in Section~\ref{sct:architecture}, while the technological implementation is described in Section~\ref{sct:implementation}. Experimental results are described in Section~\ref{sct:experimentalresults} and conclusions are finally drawn in Section~\ref{sct:conclusions}.
\section{Background}
\label{sct:background}
The homomorphic scheme encryption is a special type of encryption that allows (a set of) operations to be performed on encrypted data, i.e., directly on the ciphertexts. More specifically, as detailed in~\cite{boemer2019ngraph}, an encryption function $E$ and its decryption function $D$ are homomorphic w.r.t. a class of functions $\mathcal{F}$ if, for any function $f \in \mathcal{F}$, we can construct a function $g$ such that $f(x) = D(g(E(x)))$ for a set of input $x$.

The HE scheme considered in this paper is the Brakerski/Fan-Vercauteren (BFV) scheme~\cite{fan2012somewhat} that, similarly to other works~\cite{cheon2017homomorphic},~\cite{brakerski2014leveled}, is based on the Ring-Learning With Errors (RLWE) problem. While a detailed description of such a problem and its security/implementation aspects can be found in~\cite{lyubashevsky2010ideal}, we here provide a brief introduction to the main concepts. The BFV scheme relies on the following set of encryption parameters (from now on denoted with $\Theta$):
\begin{itemize}
    \item $m$: Polynomial modulus degree,
    \item $p$: Plaintext modulus, and
    \item $q$: Ciphertext coefficient modulus.
\end{itemize}
The parameter $m$  must be a positive power of 2 and represents the degree of the cyclotomic polynomial $\Phi_m(x)$. The plaintext modulus $p$ is a positive integer that represents the module of the coefficients of the polynomial ring $R_p=\mathbb{Z}_p[x]/ {\Phi_m(x)}$ (onto which the RLWE problem is based). Finally, the parameter $q$ is a large positive integer resulting from the product of distinct prime numbers and represents the modulo of the coefficients of the polynomial ring in the ciphertext space.
A crucial concept of a HE scheme is the Noise Budget (NB) that is an indicator related to the number of operations that can be done on a ciphertext while guaranteeing the correctness of the result. This problem (i.e., the maximum number of operations on the ciphertext) comes from the fact that, during the encryption phase, noise is added to the ciphertexts to guarantee that, being $p_1 = p_2$ two plain values  to be encrypted with the same public key, the corresponding ciphertexts $c_1$ and $c_2$ are different (i.e., $c_1 \neq c_2$). All the operations performed on the ciphertext consume a certain amount of NB (depending on the type of operation and the input): operations like additions and multiplications between ciphertext and plaintext consume a small amount of NB, while multiplications between ciphertexts are particularly demanding in terms of NB. When the NB decreases to $0$, decrypting that ciphertext will produce an incorrect result.

From a practical point of view, the choice of the encryption parameters $\Theta$ determines several aspects: the initial value of the NB, its consumption during computations (hence the number of operations to be performed on a ciphertext), the level of security against ciphertext attacks, the computational load and memory occupation of the HE processing and the accuracy of the results (i.e., measuring the correctness of the decrypted values).  For example, the initial NB increases with $m$ at the expense of larger memory occupation and computational loads. 
The plaintext modulus $p$ is directly related to the accuracy of the HE processing. Despite being a very difficult parameter to be tuned, the theory states that larger values of $p$ will produce more accurate results at the expense of larger reductions of the NB. Finally, the parameter $q$ influences both the initial NB and the level of security of the encryption. A detailed description of the parameters and their effect on the HE scheme can be found in~\cite{sealmanual}.

We emphasize that choosing the best parameter configuration is a trade-off between accuracy and performance and depends on the type and complexity of the processing, the set of feasible operations and the available computational resources. Practical guidelines to choose $\Theta$ will be given in Section~\ref{subsct:encryptParamsArchitecture}.
\section{Related Literature}
\label{sct:literature}
The idea of using HE to preserve the privacy of data during the computation has been introduced in~\cite{rivest1978data}. In this work, \emph{privacy homomorphisms} are defined as encryption functions that allow one to operate on encrypted data without preliminarily decrypting the operands~\cite{rivest1978data}. The first HE schemes allow only additions~\cite{naccache1998new, okamoto1998new,paillier1999public}, or multiplications~\cite{elgamal1985public}.

The first homomorphic encryption scheme allowing both multiplication and additions has been proposed in~\cite{gentry2009fully}. There, the idea was to rely on ideal lattice-based cryptography to provide a scheme supporting additions and multiplications with theoretically-grounded security guarantees. After that,~\cite{van2010fully} extended this work  by relaxing the ideal lattice assumption (and its security), but allowing the usage of integer polynomial rings to define the cyphertexts.~\cite{brakerski2014leveled} introduces the  Brakerski-Gentry-Vaikuntanathan (BGV) scheme that relies on polynomial rings to define the cyphertexts and on the learning with error (LWE) and ring learning with errors (RLWE) problems to provide theoretically-grounded security guarantees. The RLWE problem is also the basis of the Brakerski/Fan-Vercauteren (BFV) scheme~\cite{fan2012somewhat}, detailed in Section~\ref{sct:background}, and the Cheon-Kim-Kim-Song (CKKS) scheme~\cite{cheon2017homomorphic}, that extends the polynomial rings to the complex numbers and isometric rings.

The HE schemes mentioned above are theoretical and, to be applied, have then been implemented to specific processing chains. As regards deep learning solutions, CryptoNets~\cite{gilad2016cryptonets} relies on the HE BFV scheme to execute CNNs on encrypted inputs by introducing several possible ways of approximating the non-linear computation characterizing many layers of a CNN. Similarly,~\cite{bourse2018fast} provides a fast HE scheme for the (discretized) CNN inference.
Recently, the nGraph-HE framework~\cite{boemer2019ngraph2} has been proposed. This framework allows to train CNNs in plaintext on a given hardware and deploy trained models to HE cryptosystems operating on encrypted data. 
Unfortunately, these works are specific of a given DL solution (e.g., CNNs in~\cite{gilad2016cryptonets}), whereas our architecture is meant to be general-purpose and able to hide the complexity of adopting HE solutions, similarly to what proposed in~\cite{boemer2019ngraph2}, still maintaining the \textit{as-a-service} paradigm.

The literature presents also works aiming at offering encrypted computation. For example,~\cite{yao1982protocols} proposed the Secure Multi-Party Computation (SMC) approach, where more than one actor (namely, a party) collaborate in computing a function and having only partial knowledge of the data they are working on. These solutions do not encompass HE.~\cite{barni2006privacy} applied SMC with the Pailler HE~\cite{paillier1999public} to CNNs, where a party owns the data and another owns the CNN. Hence, both the data and CNN are kept secret during the computation. Other examples can be found in~\cite{mohassel2017secureml,rouhani2018deepsecure}.
Finally, the Gazelle framework~\cite{juvekar2018gazelle} relies on SMC and HE, to provide low-latency inference for CNN.
\section{The proposed architecture}
\label{sct:architecture}
The proposed privacy-preserving distributed architecture for deep-learning-as-a-service, called \textit{HE-DL},  is shown in Figure~\ref{fig:hl_arch}. More specifically,  \textit{HE-DL} relies on a distributed approach where the \textit{Encryption} $E\left(I, \Theta, k_p\right)$ of user data $I$ and the \textit{Decryption} $D\left(\hat{y}, \Theta, k_s\right)$ of processed data $\varphi_\Theta(\hat{I})$ are carried out on the user device given the HE parameters $\Theta$ and with the public key $k_p$ and secret key $k_s$. Both $E\left(\cdot\right)$ and $D\left(\cdot\right)$ are based on HE-BFV scheme described in Section~\ref{sct:background}. 

Conversely, the deep learning processing $\varphi_\Theta(\cdot)$ is carried out in the Cloud. This is a crucial step since deep learning processing is typically highly demanding in terms of computational load and memory occupation. We emphasize that, as commented in Section~\ref{sct:background}, the considered deep-learning-as-a-service computation has to be approximated by using only addition and multiplication in order to process the ciphertext $\hat{I}$. For this reason, the set of deep-learning models \textit{DL models} $f\left(\cdot\right)$s that are made available by the Cloud are approximated through addition and multiplication, i.e., defining the set of approximated  \textit{DL models} $\varphi\left(\cdot\right)$s. 
Once approximated, $\varphi\left(\cdot\right)$s have to be encoded following the rule of the HE-BFV scheme to get the \textit{encoded deep-learning-as-a-service} $\varphi_\Theta(\cdot)$ by relying on the HE parameters $\Theta$. This encoding phase converts plain values parameters of \textit{DL models} in a form which can be computed by the HE-BFV scheme on encrypted inputs $\hat{I}$. 

The \textit{DL models} considered in this work are the CNNs aiming at classifying the input images $I$ into a class $y \in Y$. In such a scenario the proposed \textit{HE-DL} makes available the deep-learning-as-a-service computing paradigm into two different modalities:
\begin{itemize}
    \item \textit{recall}: the processing $\varphi_\Theta(\cdot)$ provides the encrypted version $\hat{y}$ of the final classification $y$ of $I$;
    \item \textit{transfer learning}: the processing $\varphi_\Theta(\cdot)$ provides the encrypted version of a processing stage of the considered CNN applied to the input image $I$. The final classification $y$ is carried out on the User Device thanks to a suitably-trained classifier (e.g., a Support Vector Machine or a neural based classifier).
\end{itemize}
These two modalities will be detailed in the rest of the section, together with the description of the encryption/decryption phases, the approximation and encoding of CNNs, the configuration of the encryption parameters and the communication between user device and Cloud.
\subsection{Encryption and Decryption}
Let $\mathcal{P}$ be a process generating images $I \in{\mathbb{R}^{w \times h \times c}}$ of height $h$, width $w$ and channels $c$ and let $\Theta=\left\{m, p, q\right\}$ be the array of encryptions parameters, as defined in Section~\ref{sct:background}. 

The \textit{encryption} function $E\left(I, \Theta, k_p\right)$ transforms (based on the HE-BFV scheme) a plain image $I$ into an encrypted image $\hat{I}$ given the HE encryption parameters $\Theta$ with the support of a public key $k_p$.  
The \textit{decryption} function $D(\hat{y}, \Theta, k_s)$ operates on the encrypted output $\hat{y}$ of the computation $\varphi_\Theta(\hat{I})$, being $\hat{I}$ the encrypted image.  More specifically, $D(\hat{y}, \Theta, k_s)$ computes the plain output $y$ given the same set of parameters $\Theta$ and the secret key $k_s$ (corresponding to $k_p$). The semantic of  $y$ depends on the considered working modality of \textit{HE-DL}:
\begin{itemize}
    \item $y$ is  the classification label of the  input image $I$ in the \textit{recall}  modality;
    \item $y$ is an array of extracted features representing the values of the activation function of a given layer of the CNN in the \textit{transfer learning} modality. 
\end{itemize}
\subsection{Approximated and encoded DL processing}
\label{subsct:cnnProcessing}
We emphasize that the proposed architecture \textit{HE-DL} is general enough to employ a wide range of machine/deep learning models. In this paper, we decided to focus on CNNs for two main reasons. First, CNNs are widely-used and very-effective solutions for image classifications. Second, for most of their processing, CNNs are composed of addition and multiplication operations making them suitable candidates to be considered within a HE scheme.

Let $f\left(\mathcal{I}\right)$ be a CNN composed of $L$ layers $\eta^{(l)}_{\theta_l}$ with parameters $\theta_l$ and $l=1,\ldots,L$, aimed at extracting features and  providing the classification output $y$ of an input image $\mathcal{I}$.  The general architecture of $f\left(\mathcal{I}\right)$ is shown in Figure~\ref{fig:classicCNN}.

As mentioned above, in order to be used with HE, CNNs have to be approximated by considering only computing layers and activation functions that are suitable for the considered HE-BFV scheme. Given that only addition and multiplication are permitted, only polynomials functions can be computed directly, while non-polynomials operations must be either approximated with a polynomial form or replaced with other (and permitted) types of operations. For instance, the ReLU activation function is a non-polynomial operation, hence it cannot be considered in the HE scenario.
Similarly to what done in~\cite{gilad2016cryptonets}, in the proposed \textit{HE-DL} architecture, we define the approximated CNN model $\varphi\left(\cdot\right)$ of the original CNN $f\left(\cdot\right)$ by considering the following rules:
\begin{itemize}
    \item  the ReLU activation function is replaced with a Square activation function that simply squares the input value;
    \item  the max-Pooling operator is replaced with the average one, with the division converted to a  multiplication by $\frac{1   }{f_s}$, where $f_s$ is the pooling size (fixed and a-priori known).
    \item Approximate the other non-polynomial layers as in~\cite{gilad2016cryptonets}.
\end{itemize}
\begin{figure}
    \centering
    \includegraphics[width=\linewidth, height=8cm]{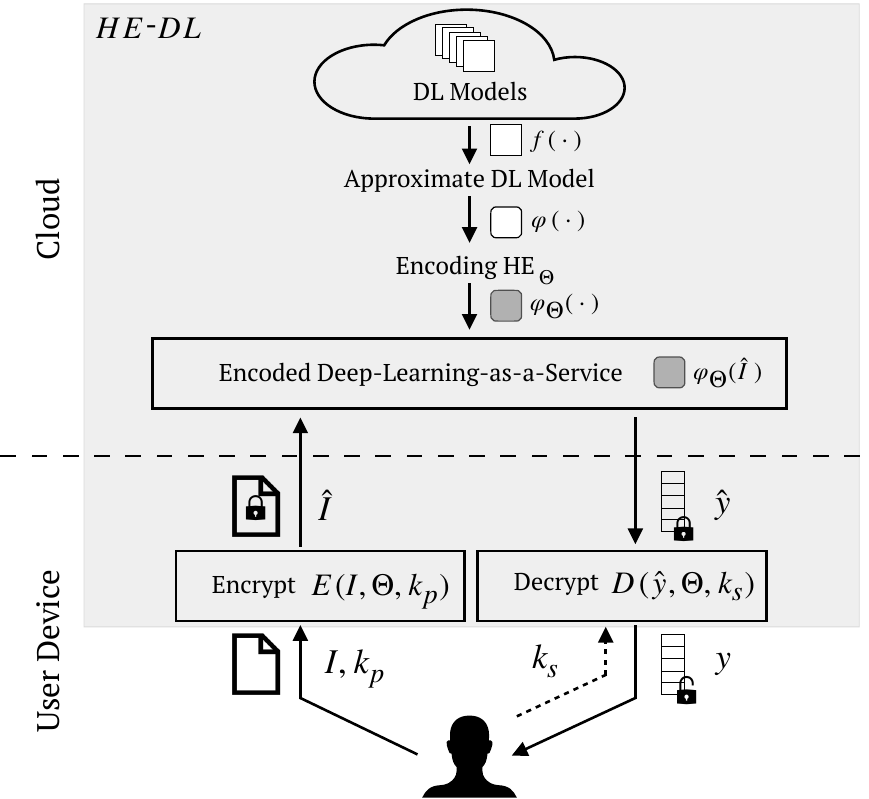}
    \caption{The proposed privacy-preserving architecture for  deep-learning-as-a-service.}
    \vspace*{-0.5cm}
    \label{fig:hl_arch}
\end{figure}
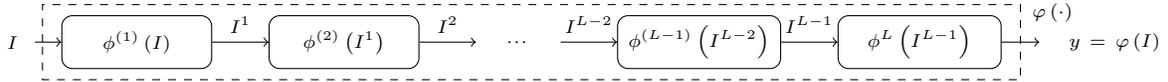
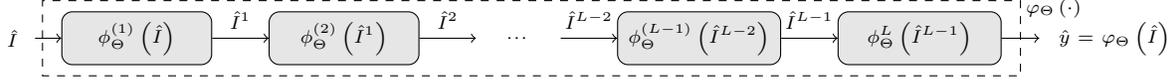
\begin{figure*}[!t]
	\centering
	%
	%
	\subfloat[][The plain processing of an approximated CNN $\varphi\left(\cdot\right)$ composed of $L $ layers. Each layer $\phi^l$, with $1\le l\le L$ is here composed only of multiplications and/or additions. The difference w.r.t. the \textit{usual} CNN-based classifier $f\left(\cdot\right)$ relies only in these approximations. Note that also the layers parameters $\theta_l$s, here omitted, may require to be approximated (and referred to as $\hat{\theta_l}$) in the approximated CNN $\varphi\left(\cdot\right)$ .]
	{
		\begin{tikzpicture}[node distance = 2 cm, auto, align=center]%
		\centering
		\node (input) [tblock, text width = 1em]
		{\centering $I$};
		\node (cnn1) [block, text width=5em, right=0.35cm of input.0]
		{\centering $\phi^{(1)}\left(I\right)$};
        \node (cnn2) [block, text width=5em, right=0.75cm of cnn1]
        {\centering $\phi^{(2)}\left(I^1\right)$};
        \node (dots) [tblock, text width=2.5em, right=0.75cm of cnn2]
        {\centering \textellipsis};
        \node (cnnx) [block, text width=5.5em, right=0.75cm of dots]
        {\centering $\phi^{(L-1)}\left(I^{L-2}\right)$};
        \node (cnnl) [block, text width=5.5em, right=0.75cm of cnnx]
        {\centering $\phi^L\left(I^{L-1}\right)$};
        \node (output) [tblock, text width=5em, right=0.5cm of cnnl]
        {$y = \varphi \left(I\right)$};
		%
		%
		%
		\draw[->] (input) -- (cnn1);
		\draw[->] (cnn1) -- 
		node [tblock, text width=2em, xshift=0.0cm, yshift=0.0cm] {$I^1$}
		(cnn2);
		\draw[->] (cnn2) -- 
		node [tblock, text width=2em, xshift=0.0cm, yshift=0.0cm] {$I^2$}
		(dots);
		\draw [->] (dots) --
		node [tblock, text width=2em, xshift=0.0cm, yshift=0.0cm] {$I^{L-2}$}
		(cnnx);
		\draw[->] (cnnx) -- 
		node [tblock, text width=2em, xshift=0.0cm, yshift=0.0cm] {$I^{L-1}$}
		(cnnl);
		\draw[->] (cnnl) -- (output);
		%
		\draw[-, dashed] ($(cnn1.180)+(-0.25,0)$) -- ++(0,0.5) -| 
		node [tblock, text width=2em, xshift=-0.05cm, yshift=-0.1cm] {$\varphi \left(\cdot\right)$}
		($(cnnl.0)+(0.25,0)$) -- ++(0,-0.5) -|  ($(cnn1.180)+(-0.25,0)$);
		\end{tikzpicture}
		\label{fig:classicCNN}
	}

	\subfloat[][The encrypted processing of the CNN $\varphi_\Theta$$\left(\cdot\right)$. The CNN is encoded with HE parameters $\Theta$, operates on images $\hat{I}$s with the same parameters $\Theta$ and returns the encrypted classification output $\hat{y}$.]
	{
		\begin{tikzpicture}[node distance = 2 cm, auto, align=center]%
		\centering
		\node (input) [tblock, text width = 1em]
		{\centering $\hat{I}$};
		\node (cnn1) [block, text width=5em, right=0.35cm of input.0, fill=gray!20]
        {\centering $\phi_\Theta^{(1)}\left(\hat{I}\right)$};
        \node (cnn2) [block, text width=5em, right=0.75cm of cnn1, fill=gray!20]
        {\centering $\phi_\Theta^{(2)}\left(\hat{I}^1\right)$};
        \node (dots) [tblock, text width=2.5em, right=0.75cm of cnn2]
        {\centering \textellipsis};
        \node (cnnx) [block, text width=5.5em, right=0.75cm of dots, fill=gray!20]
        {\centering $\phi_\Theta^{(L-1)}\left(\hat{I}^{L-2}\right)$};
        \node (cnnl) [block, text width=5.5em, right=0.75cm of cnnx, fill=gray!20]
        {\centering $\phi_\Theta^L\left(\hat{I}^{L-1}\right)$};
        \node (output) [tblock, text width=5em, right=0.5cm of cnnl]
        {$\hat{y} = \varphi_\Theta \left(\hat{I}\right)$};
		%
		
		%
		%
		%
		\draw[->] (input) -- (cnn1);
        \draw[->] (cnn1) -- 
        node [tblock, text width=2em, xshift=0.0cm, yshift=0.0cm] {$\hat{I}^1$}
        (cnn2);
        \draw[->] (cnn2) -- 
        node [tblock, text width=2em, xshift=0.0cm, yshift=0.0cm] {$\hat{I}^2$}
        (dots);
        \draw [->] (dots) --
        node [tblock, text width=2em, xshift=0.0cm, yshift=0.0cm] {$\hat{I}^{L-2}$}
        (cnnx);
        \draw[->] (cnnx) -- 
        node [tblock, text width=2em, xshift=0.0cm, yshift=0.0cm] {$\hat{I}^{L-1}$}
        (cnnl);
        \draw[->] (cnnl) -- (output);
        %
        \draw[-, dashed] ($(cnn1.180)+(-0.25,0)$) -- ++(0,0.5) -| 
        node [tblock, text width=2em, xshift=-0.05cm, yshift=-0.1cm] {$\varphi_\Theta\left(\cdot\right)$}
        ($(cnnl.0)+(0.25,0)$) -- ++(0,-0.5) -|  ($(cnn1.180)+(-0.25,0)$);
		\end{tikzpicture}
		\label{fig:encryptedCNN}
	}
	\caption{A comparison of the plain and approximated CNN processing with the encrypted one. The layers' parameters $\theta_l$s are omitted to simplify the notation.} 
    \vspace*{-0.2cm}
	\label{fig:architectureComparison}
\end{figure*}
The result of this approximation is a CNN $\varphi\left(\cdot\right)$ whose processing layers $\phi^{(l)}_{\hat{\theta_l}}$ can be encoded with the considered HE-BFV  scheme. To simplify the notation, the parameters of each layer $\theta_l$ or $\hat{\theta_l}$ are omitted from now on. 
It is important to note that, after performing the replacement of the non-polynomial layers, the model has to be trained again. This is necessary because the weights of the plain model are not valid anymore if the activation functions or other layers have been replaced by different ones. Hence, to provide a  deep-learning-as-a-service,  $\varphi\left(\cdot\right)$ must be retrained with the same settings in which the plain one was trained (e.g., same dataset, same learning algorithm, etc..). Obviously, if the original model $f\left(\cdot\right)$ already contains  HE-compatible processing layers, this procedure is not necessary.
Moreover, it's noteworthy that this approximation process can introduce a variation in the accuracy between $f\left(\cdot\right)$ and $\varphi\left(\cdot\right)$. This aspect will be explored in the experimental section described in Section~\ref{sct:experimentalresults}. We emphasize that we considered (and made available to the scientific community) two already approximated and trained models, i.e., a 5-layers CNN and a 6-layers CNN trained on the FashionMNIST data-set~\cite{xiao2017fashion}; these models will be used in the experimental section.

To work with the encrypted images $\hat{I}$s, the suitably approximated CNN $\varphi$ must be encoded with the parameters $\Theta$ as defined by the HE-BFV  scheme leading to the encoded CNN $\varphi_\Theta(\cdot)$.  As shown in Figure~\ref{fig:encryptedCNN}, the HE-based \textit{encrypted processing} can be formalized as follows: 
\begin{equation}\label{eq:y}
y =
D\left(
\hat{y}, \Theta, k_s
\right)
=
D\left(
\varphi_\Theta \left(
E\left(
I, \Theta, k_p
\right)
\right), \Theta, k_s
\right), 
\end{equation}
where $\hat{y}$ represents the image $I$'s encrypted classification.
%
\subsection{DL models: recall and transfer learning}
\label{subsct:recallandtransferlearning}
As mentioned above, the deep-learning-as-a-service computing paradigm is made available in two different modalities, \textit{recall} and \textit{transfer learning}. The difference between the two modalities lies in how Eq.~\eqref{eq:y} is implemented. The former operates on the decrypted output $y$ of the CNN $\varphi$ last layer $L$ (typically a softmax on top of a classification layer), whereas the latter one works on the features $I^{\tilde{l}}$ extracted at a given CNN level $\tilde{l}$, with $1 \le \tilde{l} < L$ (typically a convolutional or pooling one). The two modalities are detailed in what follows. 
\subsubsection*{Recall}
This is the modality where the user relies on one of the ready-to-use encoded CNN $\varphi_\Theta(\cdot)$s to classify the image $I$. More precisely, the user wants the image $I$ to be encrypted into $\hat{I}$ and to be forwarded through all the layers of the encoded CNN $\varphi_\Theta$, hence obtaining the final result $\hat{y}$ of the classification task, without transmitting the image $I$ to the service provider.
The assumption underlying this modality is that the chosen model $\varphi_\Theta(\cdot)$ is trained to classify images of the same domain of the input image $I$ (e.g., the model $\varphi_\Theta(\cdot)$ is trained to recognize the digits and $I$ is an image of a digit).
\subsubsection*{Transfer Learning}
When the application problem of the user is not matched by the model $\varphi_\Theta(\cdot)$s (e.g., the user wants to distinguish between cars and bikes while available models have been trained to classify digits or faces), the transfer learning modality comes into play. In fact, following the transfer learning paradigm~\cite{yosinski2014transferable,alippi2018moving},  the processing of a pre-trained CNN can be split into two parts: feature extraction and classification. The feature extraction processing represents a pre-trained feature extractor able to feed an ad-hoc classifier trained on the specific image classification problem (that can be different from the one originally used to train the CNN). This allows to use part of a pre-trained CNN and train only a final classifier (hence reducing the complexity for the training and the number of images required for the training.

In our scenario, the encrypted input images, $\hat{I}$s, will be forwarded through the encoded model $\varphi_\Theta$ up to a layer $\tilde{l}$. More specifically, $\varphi_\Theta$ comprises layers from $1$ to 
$\tilde{l}$, with $1\le\tilde{l}\le L$, whereas all the (eventually) remaining layers, from $\tilde{l}+1$ to the final one $L$ remain plain and operate on the decrypted output of layer $\tilde{l}$, i.e., 
$I^{\tilde{l}} = D\left(
\varphi_\Theta^{\tilde{l}} \left(
E\left(
I, \Theta, k_p
\right)
\right), \Theta, k_s
\right)$, where $\varphi_\Theta^{\tilde{l}}$ represents the encoded CNN up to layer $\tilde{l}$ with parameters $\Theta$. The output of the model will be, in this case, the features extracted from every image $I$.  The user may use these features to train a local classifier (e.g., a Support Vector Machine); an example will be shown in section~\ref{sct:experimentalresults}. 

We emphasize that, following such an approach, the user is able to locally train a classifier on the decrypted vectors $y=D\left(\hat{y}, \Theta, k_s\right)$, being $\hat{y}$ the output of $\varphi_{\Theta}$. A set of $K$ images $\{I_1,\ldots,I_K\}$ is sent to \textit{HE-DL} providing the corresponding output $\{\hat{y}_1,\ldots,\hat{y}_K\}$ that are locally decrypted into $\{y_1,\ldots,y_K\}$. The vector set $\{y_1,\ldots,y_K\}$ is used together with the corresponding labels (that are available to the user) to locally train a classifier. Once trained, the system is ready-to-use: the user can send an encrypted image $\hat{I}$ to the Cloud, receive the CNN output $\hat{y}$, decrypt it to $y$ and apply the classifier on $y$.
\subsection{Encryption parameters}
\label{subsct:encryptParamsArchitecture}
As already mentioned, the choice of $\Theta$ is critical to get correct processing of the encrypted image $\hat{I}$.
The choice for $q$ is particularly difficult and influences the security of the scheme. For this purpose SEAL library~\cite{sealcrypto} provides a specific function that, given the polynomial modulus degree $m$ and the desired AES-equivalent security level ($sec$), returns a suggested value for $q$~\cite{sealmanual}.
In this work we considered  \textit{sec} equals to $128$ \textit{bits} which is the default value of SEAL.
Hence, we relied on the SEAL function to automatically set the values of $q$ to guarantee a $128$ \textit{bits} security level, while we selected $m \in \{1024, 2048, 4096\}$ and $p \in \{32, 712, 37780, 1.3\cdot 10^5, 1.5\cdot 10^5, 2.6\cdot 10^5, 5.2\cdot 10^5, 6.0\cdot 10^5, 2.1\cdot 10^6, 1.3\cdot 10^8, 1.5\cdot 10^8\}$, through an experimental analysis.
The effects of the different choices for $\Theta$ are shown in Section~\ref{sct:experimentalresults}.
\subsection{Communication between User Device and Cloud}
The communication between the User Device and the Cloud is carried out through a JSON-format message. More specifically, being an on-demand computation, clients have to perform a request to the on-line deep-learning-as-service provider including:
\begin{itemize}
    \item a set of parameters, including the encryption parameters $m$ and $p$, the security level $sec$, the identifier of the chosen DL model $\varphi_\Theta(\cdot)$ to use in the computation, and the specific layers to use (which will determine the modality, i.e., recall or transfer learning);
    \item the encrypted image $\hat{I}$ on which the computation is performed, which has to be encrypted using a public key generated according to the encryption parameters.
\end{itemize}
Information about the available models will be published by the provider.
$\hat{I}$ is transmitted as a vector in which the ciphertexts are encoded as base64 strings making it possible to embed them into JSON files.
Once the computation has been carried out, the Cloud responds with a JSON message containing the encrypted result vector.

As an example, if the user wants to classify a batch of 20 images from the FashionMNIST using \emph{Model1}, the JSON will contain $[m=2048, p=600201, sec=128]$, the details of the models ("model"="Model1", "layers"= $[0, 1, 2, 3, 4, 5, 6]$) and the encrypted image (a vector with dimensions $[20, 1, 28, 28]$). The answer JSON message will contain the encrypted classification, so a vector of dimension $[20, 10]$ of ciphertexts.  

\section{Implementation}
\label{sct:implementation}
%
%
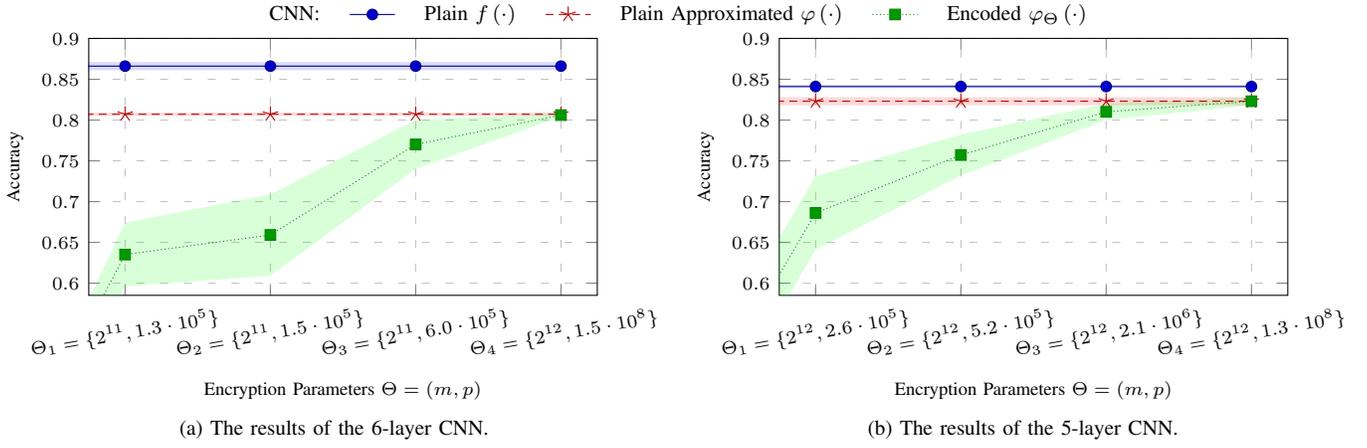
\begin{figure*}[!t]
	\centering
	%
	%
	\begin{tikzpicture}
	\begin{customlegend}[legend columns=8,legend style={align=left,draw=none,column sep=2ex,font=\footnotesize},legend entries={CNN:,Plain $f\left(\cdot\right)$,Plain Approximated $\varphi\left(\cdot\right)$, Encoded $\varphi_{\Theta}\left(\cdot\right)$,},cycle list name=custom]
	%
	%
	\addlegendimage{empty legend}
	\addlegendimage{blue!50!black,every mark/.append style={fill=blue!80!black},mark=*}   
	\addlegendimage{red!70!black,densely dashed,every mark/.append style={solid, fill=red!80!black},mark=star,mark options={scale=2}}
	\addlegendimage{green!50!black,densely dotted, mark=square*,every mark/.append style={solid,fill=green!60!black}}
	\end{customlegend}
	\end{tikzpicture}
	\vspace*{-0.5cm}

	\subfloat[][The results of the 6-layer CNN.]
	{
		\begin{tikzpicture}
		\begin{axis}[
		width=0.46\textwidth,
		height=5cm,
		xmin=1.75,
		xmax=5.25,
		xtick={2,3,4,5},
		xticklabels = {{$\Theta_1=\{ 2^{11}, 1.3\cdot 10^5\}$}, {$\Theta_2=\{ 2^{11}, 1.5\cdot 10^5 \}$}, {$\Theta_3=\{2^{11}, 6.0\cdot 10^5\}$}, { $\Theta_4=\{2^{12}, 1.5\cdot 10^8\}$}},
        xticklabel style={rotate=10},
        ymin=0.585,
     	ymax=0.90,
        ytick={0.10,0.15,...,1.025},
		extra y ticks={0,0.05},
		extra y tick labels={0,0.05},
		xmajorgrids,
		ymajorgrids,
		grid style={loosely dashed,ultra thin},
		xlabel={Encryption Parameters $\Theta=(m,p) $},
		ylabel=Accuracy,
		mark options={scale=1},
		cycle list name=custom,
		]
		\addplot
		table [x=theta,y=pma,col sep=comma] 
		{csv/recall_model_1.csv};
		\addplot[name path=top,draw=none,forget plot]
		table [x=theta,y expr=\thisrow{pma}+\thisrow{pmstd},col sep=comma] 
		{csv/recall_model_1.csv};
		\addplot[name path=bot,draw=none,forget plot]
    	table [x=theta,y expr=\thisrow{pma}-\thisrow{pmstd},col sep=comma] 
		{csv/recall_model_1.csv};
		\addplot [forget plot, draw=none,opacity=0.15,fill=blue]
		fill between[of=top and bot];
		\addplot
		table [x=theta,y=ama,col sep=comma] 
		{csv/recall_model_1.csv};
		\addplot[name path=top,draw=none,forget plot]
		table [x=theta,y expr=\thisrow{ama}+\thisrow{amstd},col sep=comma] 
		{csv/recall_model_1.csv};
		\addplot[name path=bot,draw=none,forget plot]
		table [x=theta,y expr=\thisrow{ama}-\thisrow{amstd},col sep=comma] 
		{csv/recall_model_1.csv};
		\addplot [forget plot, draw=none,opacity=0.15,fill=red]
		fill between[of=top and bot];
		\addplot
		table [x=theta,y=thetaa,col sep=comma] 
		{csv/recall_model_1.csv};
		\addplot[name path=top,draw=none,forget plot]
		table [x=theta,y expr=\thisrow{thetaa}+\thisrow{thetastd},col sep=comma] 
		{csv/recall_model_1.csv};
		\addplot[name path=bot,draw=none,forget plot]
		table [x=theta,y expr=\thisrow{thetaa}-\thisrow{thetastd},col sep=comma] 
		{csv/recall_model_1.csv};
		\addplot [forget plot, draw=none,opacity=0.15,fill=green]
		fill between[of=top and bot];
		\end{axis}
		\end{tikzpicture}
		\label{fig:resultsFashionMNISTmodel1}
	}
	\subfloat[][The results of the 5-layer CNN.]
	{
		\begin{tikzpicture}
		\begin{axis}[
		width=0.46\textwidth,
		height=5cm,
		xmin=2.75,
		xmax=6.25,
		xtick={3,4,5,6},
		xticklabels = {{$\Theta_1=\{2^{12}, 2.6\cdot 10^5\}$}, {$\Theta_2=\{2^{12}, 5.2\cdot 10^5\}$}, { $\Theta_3=\{2^{12}, 2.1\cdot 10^6\}$}, {$\Theta_4=\{2^{12}, 1.3\cdot 10^8\}$}},
        xticklabel style={rotate=10},
		ymin=0.585,
		ymax=0.90,
		ytick={0.10,0.15,...,1.025},
		extra y ticks={0,0.05},
		extra y tick labels={0,0.05},
		xmajorgrids,
		ymajorgrids,
		grid style={loosely dashed,ultra thin},
		xlabel={Encryption Parameters $\Theta=(m,p) $},
		ylabel=Accuracy,
		mark options={scale=1},
		cycle list name=custom,
		]
		\addplot
		table [x=theta,y=pma,col sep=comma] 
		{csv/recall_model_2.csv};
		\addplot[name path=top,draw=none,forget plot]
		table [x=theta,y expr=\thisrow{pma}+\thisrow{pmstd},col sep=comma] 
		{csv/recall_model_2.csv};
		\addplot[name path=bot,draw=none,forget plot]
		table [x=theta,y expr=\thisrow{pma}-\thisrow{pmstd},col sep=comma] 
		{csv/recall_model_2.csv};
		\addplot [forget plot, draw=none,opacity=0.15,fill=blue]
		fill between[of=top and bot];
		\addplot
		table [x=theta,y=ama,col sep=comma] 
		{csv/recall_model_2.csv};
		\addplot[name path=top,draw=none,forget plot]
		table [x=theta,y expr=\thisrow{ama}+\thisrow{amstd},col sep=comma] 
		{csv/recall_model_2.csv};
		\addplot[name path=bot,draw=none,forget plot]
		table [x=theta,y expr=\thisrow{ama}-\thisrow{amstd},col sep=comma] 
		{csv/recall_model_2.csv};
		\addplot [forget plot, draw=none,opacity=0.15,fill=red]
		fill between[of=top and bot];
		\addplot
		table [x=theta,y=thetaa,col sep=comma] 
		{csv/recall_model_2.csv};
		\addplot[name path=top,draw=none,forget plot]
		table [x=theta,y expr=\thisrow{thetaa}+\thisrow{thetastd},col sep=comma] 
		{csv/recall_model_2.csv};
		\addplot[name path=bot,draw=none,forget plot]
		table [x=theta,y expr=\thisrow{thetaa}-\thisrow{thetastd},col sep=comma] 
		{csv/recall_model_2.csv};
		\addplot [forget plot, draw=none,opacity=0.15,fill=green]
		fill between[of=top and bot];
		\end{axis}
		\end{tikzpicture}
		\label{fig:resultsFashionMNISTmodel2}
	}

	\caption{The \textit{recall} accuracy results of both the 6-layer CNN and the 5-layer CNN on the FashionMNIST dataset~\cite{xiao2017fashion}, with the standard deviation over five experiments. 
     For each considered encryption parameters $\Theta_i$, three cases are compared: the plain CNN without approximations $f\left(\cdot\right)$, the same plain CNN approximated to have only additions and multiplications $\varphi\left(\cdot\right)$, and, finally, the encoded CNN with $\Theta_i$, i.e., $\varphi_{\Theta_i}\left(\cdot\right)$.  It is noteworthy to point out that with encryption parameters $\Theta_i$ \textit{smaller} (i.e., smaller $m$ and $p$) than those shown, the accuracy quickly drops to that of a random classifier.}
	\label{fig:resultsFashionMNIST}
    \vspace*{-0.2cm}
\end{figure*}
The architecture introduced in the previous section has been implemented through a Python library, named PyCrCNN, comprising a client-side and a server application. PyCrCNN supports the encryption and decryption of batches of integer or float values and the application of the common layers used in CNNs like convolutional layers, average pool layers, and fully connected layers, relying on PyTorch library~\cite{paszke2019pytorch}. For the HE operations, PyCrCNN relies on the Pyfhel library v2.0.1~\cite{pyfhel}, Laurent (SAP) and Onen (EURECOM), licensed under the GNU GPL v3 license\footnote{Pyfhel is a wrapper on the Microsoft SEAL library.}.
\subsection{Client}
The client-side can encrypt the input images $I$s and decrypt the resulting answer $\hat{y}$ in a transparent way with respect to the user. Once the parameters are set (which include encryption parameters $\Theta$, name and layers of the chosen model $\varphi(\cdot)$, server URL and port), the client-side of PyCrCNN exposes a function which receives $I$ as a NumPy~\cite{walt2011numpy} vector and returns $y$ as a NumPy vector; this makes it compliant with many machine learning frameworks for Python. Before starting the computation, a public and secret key pair $(k_p, k_s)$ is generated. The input batch is encrypted and encoded in base64 strings that will be included in the JSON payload along with the parameters $\Theta$ (as described in the previous section). To perform the request, the JSON payload is uploaded to an Amazon S3 bucket. Then, a POST request containing the address to the uploaded data is made to the deep-learning-as-a-service URL and, once the reply $\hat{y}$ is received, the resulting batch is downloaded from the bucket and decrypted using the key $k_s$ generated before. Finally, the user receives back the decrypted value $y$ as a NumPy array.
\subsection{Server}
The server side of the deep-learning-as-a-service must be invoked via web API. For this purpose, we relied on a set of Amazon Web Services (AWS) tools comprising Sagemaker, Elastic Container Registry (ECR), AWS Lambda, API Gateway, and S3. More precisely, we extended the built-in models offered by Sagemaker with our own custom algorithm, i.e., PyCrCNN, by creating a Docker container compliant with Sagemaker Docker Images specifications, uploading it to ECR and deploying the model on Sagemaker. The Docker container uses NginX as a web server, Gunicorn as a WSGI and Flask, a Python library, as a web framework to expose the APIs required by Sagemaker. With a mock \emph{fit} method we load and store the model to S3; with the actual \emph{predict} method the model performs the feature extraction task. Hence, the proposed deep-learning-as-a-service is made available through a REST API: the client invokes the endpoint URL with a POST request whose JSON payload contains the S3 path to the image $\hat{I}$ encrypted by the client and the aforementioned encryption parameters $\Theta$. In order to obtain a JSON-serializable payload, we encode the encrypted image $\hat{I}$ as a base64 string. The client receives back the encrypted server response $\hat{y}$ as a base64 string containing the extracted features.
\section{Experimental Results}
\label{sct:experimentalresults}
The aim of this section is to evaluate the accuracy and the computation load of the deep-learning-as-service provided trough PyCrCNN both in recall and transfer learning modality. Section~\ref{subsct:cnnExperimental} describes the CNNs provided by the deep-learning-as-service, while Section~\ref{subsct:datasetsExperimental} details the considered datasets. Accuracy and computational load on both recall and transfer learning modality are shown in Sections~\ref{subsct:recallExperimental},~\ref{subsct:transferExperimental} and~\ref{subsct:timingExperimental}.
\subsection{Description of the CNNs}
\label{subsct:cnnExperimental}
The first deep learning model is a \textit{6-layer CNN} composed by the following processing layers: a convolutional layer with 8 3x3 filters, a 2x2 maximum pooling layer with stride 3, a convolutional layer with 16 3x3 filters and stride 2, a 2x2 maximum pooling layer and two fully-connected layers with 16 and 10 neurons respectively.
The second deep learning model is a \textit{5-layer CNN} composed by a convolutional layer with 16 3x3 filters with stride 3 and a ReLU activation function, a 3x3 maximum pooling layer with stride 3 and two fully connected layers with 72 and 10 neurons respectively.
\subsection{Datasets}
\label{subsct:datasetsExperimental}
Two datasets have been considered in the analysis: 
\begin{itemize}
    \item \textit{MNIST}~\cite{lecun1998mnist} is a datasets of handwritten digits composed of 70000 grey-scale 28x28 images, belonging to 10 classes. From the datasets, 5000 images were used for training and 5000 for validation.
    \item \textit{FashionMNIST}~\cite{xiao2017fashion} is a datasets of fashion products composed of 70000 grey-scale 28x28 images, belonging to 10 classes. From the datasets, 60000 images were used for training and 10000 for validation.
\end{itemize}
In particular, the \textit{FashionMNIST} dataset has been considered in the \textit{recall} modality, while \textit{MNIST} has been used in the \textit{transfer learning} one.
\subsection{Recall}
\label{subsct:recallExperimental}
In this modality a user wants to use a deep-learning-as-a-service model $\varphi_\Theta(\cdot)$ published by a Cloud service provider, obtaining the classification $y$ of an input image $I$.
Figure~\ref{fig:resultsFashionMNISTmodel1} and~\ref{fig:resultsFashionMNISTmodel2} show the accuracy of the 6-layers CNN and the 5-layers CNN on the FashionMNIST dataset, respectively, with respect to different values of $\Theta$ (the parameter $q$ has been omitted since automatically set). The two CNNs in both the configurations, \textit{plain} and \textit{approximated}, have been trained on the FashionMNIST training dataset for 20 epochs, with a learning rate of 0.001. 
As expected,  the accuracy of the encoded model $\varphi_\Theta(\cdot)$ increases with $m$ and $p$. In particular, the configuration of parameters $\Theta_4$ (characterized by the largest values of $m$ and $p$) provides the same performance of the approximated model $\varphi\left(\cdot\right)$ operating on plain data. 
It is noteworthy to point out that the 6-layers CNN (Figure~\ref{fig:resultsFashionMNISTmodel1}), when used plain, is indeed better than the 5-layers CNN (Figure~\ref{fig:resultsFashionMNISTmodel2}): in fact, the former has higher accuracy than the latter. However, after the approximation, the 5-layers CNN outperforms the 6-layers CNN. This suggests that the approximated CNN $\varphi\left(\cdot\right)$  could be designed from scratch.
\subsection{Transfer learning}
\label{subsct:transferExperimental}
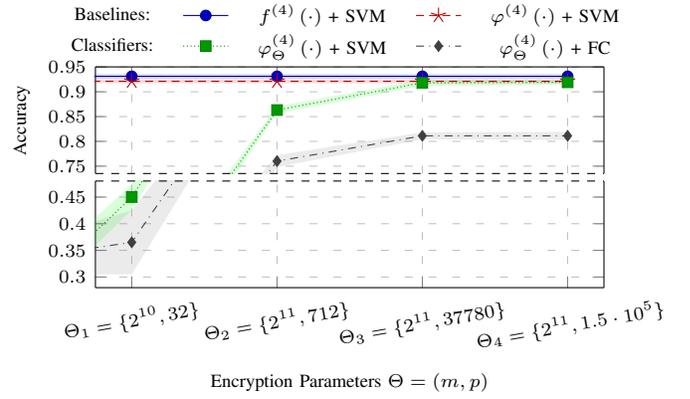
\begin{figure}[!t]
	\centering
	%
	%
	\begin{tikzpicture}
	\begin{customlegend}[legend columns=3,legend style={align=left,draw=none,column sep=2ex,font=\footnotesize},legend entries={\scriptsize Baselines:, \scriptsize $f^{(4)}\left(\cdot\right)$ + SVM,\scriptsize $\varphi^{(4)}\left(\cdot\right)$ + SVM,\scriptsize Classifiers:,\scriptsize $\varphi^{(4)}_{\Theta}\left(\cdot\right)$ + SVM,\scriptsize $\varphi^{(4)}_{\Theta}\left(\cdot\right)$ + FC},cycle list name=custom]
	%
	%
	\addlegendimage{empty legend}
	\addlegendimage{blue!50!black,every mark/.append style={fill=blue!80!black},mark=*}   
	\addlegendimage{red!70!black,densely dashed,every mark/.append style={solid, fill=red!80!black},mark=star,mark options={scale=2}}
    \addlegendimage{empty legend}
	\addlegendimage{green!50!black,densely dotted, mark=square*,every mark/.append style={solid,fill=green!60!black}}
    \addlegendimage{darkgray,dashdotted,every mark/.append style={solid,fill=darkgray},mark=diamond*}   
	\end{customlegend}
	\end{tikzpicture}
	\vspace*{-0.3cm}

        \begin{tikzpicture}
        \begin{groupplot}[
        group style={
            group size=1 by 2,
            xticklabels at=edge bottom,
            vertical sep=0.1cm,
        },
        width=0.46\textwidth,
        xmin=2.75,
        xmax=6.25,
        xtick={3,4,5,6},
        xticklabels = {{$\Theta_1=\{ 2^{10}, 32\}$}, {$\Theta_2=\{ 2^{11}, 712\}$}, {$\Theta_3=\{2^{11}, 37780\}$}, { $\Theta_4=\{2^{11}, 1.5\cdot 10^5\}$}},
        xticklabel style={rotate=10},
        xmajorgrids,
        ymajorgrids,
        grid style={loosely dashed,ultra thin},
        cycle list name=custom,
        ]
        \nextgroupplot[
        ymin=0.735,
        ymax=0.95,
        ytick={0.10,0.15,...,1.025},
        axis x line=top,
        ylabel=Accuracy,
        height=3cm]

        \addplot
        table [x=theta,y=pma,col sep=comma] 
        {csv/training_model_1.csv};
        \addplot[name path=top,draw=none,forget plot]
        table [x=theta,y expr=\thisrow{pma}+\thisrow{pmstd},col sep=comma] 
        {csv/training_model_1.csv};
        \addplot[name path=bot,draw=none,forget plot]
        table [x=theta,y expr=\thisrow{pma}-\thisrow{pmstd},col sep=comma] 
        {csv/training_model_1.csv};
        \addplot [forget plot, draw=none,opacity=0.15,fill=blue]
        fill between[of=top and bot];
        \addplot
        table [x=theta,y=ama,col sep=comma] 
        {csv/training_model_1.csv};
        \addplot[name path=top,draw=none,forget plot]
        table [x=theta,y expr=\thisrow{ama}+\thisrow{amstd},col sep=comma] 
        {csv/training_model_1.csv};
        \addplot[name path=bot,draw=none,forget plot]
        table [x=theta,y expr=\thisrow{ama}-\thisrow{amstd},col sep=comma] 
        {csv/training_model_1.csv};
        \addplot [forget plot, draw=none,opacity=0.15,fill=red]
        fill between[of=top and bot];
        \addplot
        table [x=theta,y=svma,col sep=comma] 
        {csv/training_model_1.csv};
        \addplot[name path=top,draw=none,forget plot]
        table [x=theta,y expr=\thisrow{svma}+\thisrow{svmstd},col sep=comma] 
        {csv/training_model_1.csv};
        \addplot[name path=bot,draw=none,forget plot]
        table [x=theta,y expr=\thisrow{svma}-\thisrow{svmstd},col sep=comma] 
        {csv/training_model_1.csv};
        \addplot [forget plot, draw=none,opacity=0.15,fill=green]
        fill between[of=top and bot];
        \addplot
        table [x=theta,y=lca,col sep=comma] 
        {csv/training_model_1.csv};
        \addplot[name path=top,draw=none,forget plot]
        table [x=theta,y expr=\thisrow{lca}+\thisrow{lcstd},col sep=comma] 
        {csv/training_model_1.csv};
        \addplot[name path=bot,draw=none,forget plot]
        table [x=theta,y expr=\thisrow{lca}-\thisrow{lcstd},col sep=comma] 
        {csv/training_model_1.csv};
        \addplot [forget plot, draw=none,opacity=0.15,fill=darkgray!70!white]
        fill between[of=top and bot];
        \coordinate(da) at (rel axis cs:0,0);
        \coordinate(dal) at (rel axis cs:1,0);
        \nextgroupplot[        
        ymin=0.28,
        ymax=0.48,
        ytick={0.10,0.15,...,1.025},
        axis x line=bottom, 
        height=3cm,
        xlabel={Encryption Parameters $\Theta=(m,p) $},
         ]

        \addplot
        table [x=theta,y=pma,col sep=comma] 
        {csv/training_model_1.csv};
        \addplot[name path=top,draw=none,forget plot]
        table [x=theta,y expr=\thisrow{pma}+\thisrow{pmstd},col sep=comma] 
        {csv/training_model_1.csv};
        \addplot[name path=bot,draw=none,forget plot]
        table [x=theta,y expr=\thisrow{pma}-\thisrow{pmstd},col sep=comma] 
        {csv/training_model_1.csv};
        \addplot [forget plot, draw=none,opacity=0.15,fill=blue]
        fill between[of=top and bot];
        \addplot
        table [x=theta,y=ama,col sep=comma] 
        {csv/training_model_1.csv};
        \addplot[name path=top,draw=none,forget plot]
        table [x=theta,y expr=\thisrow{ama}+\thisrow{amstd},col sep=comma] 
        {csv/training_model_1.csv};
        \addplot[name path=bot,draw=none,forget plot]
        table [x=theta,y expr=\thisrow{ama}-\thisrow{amstd},col sep=comma] 
        {csv/training_model_1.csv};
        \addplot [forget plot, draw=none,opacity=0.15,fill=red]
        fill between[of=top and bot];
        \addplot
        table [x=theta,y=svma,col sep=comma] 
        {csv/training_model_1.csv};
        \addplot[name path=top,draw=none,forget plot]
        table [x=theta,y expr=\thisrow{svma}+\thisrow{svmstd},col sep=comma] 
        {csv/training_model_1.csv};
        \addplot[name path=bot,draw=none,forget plot]
        table [x=theta,y expr=\thisrow{svma}-\thisrow{svmstd},col sep=comma] 
        {csv/training_model_1.csv};
        \addplot [forget plot,opacity=0.15,fill=green]
        fill between[of=top and bot];
        \addplot
        table [x=theta,y=lca,col sep=comma] 
        {csv/training_model_1.csv};
        \addplot[name path=top,draw=none,forget plot]
        table [x=theta,y expr=\thisrow{lca}+\thisrow{lcstd},col sep=comma] 
        {csv/training_model_1.csv};
        \addplot[name path=bot,draw=none,forget plot]
        table [x=theta,y expr=\thisrow{lca}-\thisrow{lcstd},col sep=comma] 
        {csv/training_model_1.csv};
        \addplot [forget plot,opacity=0.15,fill=darkgray!70!white]
        fill between[of=top and bot];
        %
        \coordinate(db) at (rel axis cs:0,1);
        \coordinate(dbr) at (rel axis cs:1,1);
        \end{groupplot}
        %
            \draw[dashed] (da) -- (dal);
            \draw[dashed] (db) -- (dbr);
	\end{tikzpicture}
    \vspace*{-0.75cm}
	\caption{The \textit{transfer learning} accuracy results on the features extracted at layer $l=4$ of the 6-layer CNN to the MNIST dataset~\cite{lecun1998mnist}.
     For each considered encryption parameters $\Theta_i$, four cases are compared: the plain CNN without approximations $f^{(4)}\left(\cdot\right)$ with a SVM-based classifier, the same plain CNN approximated to have only additions and multiplications $\varphi^{(4)}\left(\cdot\right)$ with the SVM, and, the encoded CNN with $\Theta_i$, i.e., $\varphi^{(4)}_{\Theta_i}\left(\cdot\right)$ with either a SVM or a Fully-Connected classifier. }
    \vspace*{-0.2cm}
	\label{fig:resultsTransferLearning}
\end{figure}

\begin{table}
	\centering
	\scriptsize
	\caption{The three described configurations results, with a common PC as a client and an Amazon EC2 instance as a server. The main result, $t$, is the time required to process an image for each scenario. The three components of $t$ are $t_c$, the time required for the local encryption/decryption, $t_t$, the time for the data transfer and $t_s$, the time required for the processing on the Cloud. The proposed values are expressed in seconds.}
	\begin{tabular}{c c|c@{\hskip 4pt}@{\hskip 4pt}c@{\hskip 4pt}@{\hskip 4pt}c|c}
		& & $t_c$ & $t_t$ & $t_s$ & $t = t_c + t_t + t_s$ \\
		\toprule[1.5pt]
		\multirow{4}{*}{\rotatebox[origin=c]{90}{\parbox{1.75cm}{\centering Recall\\ 6-layers CNN}}} 
			&$\Theta_1$&$2.2\pm0.2$&$3.7\pm0.0$&$11.8\pm0.1$&$17.7\pm0.3$\\ \cmidrule{2-6}
			&$\Theta_2$&$2.2\pm0.1$&$3.7\pm0.0$&$11.9\pm0.1$&$17.8\pm0.2$\\ \cmidrule{2-6}
			&$\Theta_3$&$2.1\pm0.1$&$3.7\pm0.0$&$11.9\pm0.0$&$17.7\pm0.1$\\ \cmidrule{2-6}
			&$\Theta_4$&$4.7\pm0.3$&$14.7\pm0.0$&$49.7\pm0.5$&$69.1\pm0.8$\\
		\midrule[1.5pt]
		\multirow{4}{*}{\rotatebox[origin=c]{90}{\parbox{1.75cm}{\centering Recall\\ 5-layers CNN}}}
			&$\Theta_1$&$5.2\pm0.0$&$14.7\pm0.0$&$26.2\pm0.3$&$46.1\pm0.3$\\ \cmidrule{2-6}
			&$\Theta_2$&$5.2\pm0.0$&$14.7\pm0.0$&$26.1\pm0.1$&$46.0\pm0.1$\\ \cmidrule{2-6}
			&$\Theta_3$&$5.2\pm0.0$&$14.7\pm0.0$&$25.8\pm0.1$&$45.7\pm0.1$\\ \cmidrule{2-6}
			&$\Theta_4$&$5.2\pm0.0$&$14.7\pm0.0$&$25.8\pm0.1$&$45.7\pm0.1$\\
		\midrule[1.5pt]
		\multirow{4}{*}{\rotatebox[origin=c]{90}{\parbox{1.75cm}{\centering Transfer Learning}}}
			&$\Theta_1$&$1.2\pm0.0$&$2.0\pm0.0$&$5.5\pm0.0$&$8.7\pm0.0$\\ \cmidrule{2-6}
			&$\Theta_2$&$2.4\pm0.1$&$3.9\pm0.0$&$11.6\pm0.1$&$17.9\pm0.2$\\ \cmidrule{2-6}
			&$\Theta_3$&$2.4\pm0.0$&$3.9\pm0.0$&$11.5\pm0.0$&$17.8\pm0.0$\\ \cmidrule{2-6}
			&$\Theta_4$&$2.4\pm0.0$&$3.9\pm0.0$&$11.5\pm0.0$&$17.8\pm0.0$\\
		\bottomrule[1.5pt]
	\end{tabular}
	\label{tab:timing}
\end{table}
In this modality, the user relies on deep-learning-as-a-service $\varphi_\Theta(\cdot)$  as a feature extractor to train a local classifier, as described in Section~\ref{subsct:recallandtransferlearning}. Two types of classifiers have been used, i.e., an SVM-based classifier and a Fully-Connected based classifier. Both classifiers have been trained using the features extracted from images coming from the MNIST~\cite{lecun1998mnist} dataset, using the first 4 layers of the pre-trained 6-layers CNN.  In particular, 5000 images were used for the training of the classifiers and 5000 for the testing. Figure ~\ref{fig:resultsTransferLearning} shows the accuracy of the SVM-based classifier and Fully-Connected classifier.  Different values for $\Theta$ show the impact on the precision of the extracted features, hence the accuracy of the trained classifiers. Here, two main comments arise.
First, moving from $\Theta_3$ to $\Theta_4$ (with a relevant increase in the parameter $p$) does not induce a significant improvement in the accuracy. This means that the value $p=37780$ well characterizes the processing chain of $\varphi_\Theta(\cdot)$. Secondly, $\Theta_2$ for the 6-layers CNN in the recall scenario is equal to $\Theta_4$ in the transfer learning scenario. However, in the latter case, this set of parameters provides enough NB and precision to carry out the computations correctly, while in the former case it does not. This can be explained by the fact that in this transfer learning scenario the number of encoded layers is lower than in the recall one.
\subsection{Timing}
\label{subsct:timingExperimental}
In addition to the accuracy, we evaluated the performance of the proposed PyCrCNN implementation by measuring the computational times on the client and the server-side and by estimating the transmission times of exchange information. For this purpose, we considered a single image taken from the FashionMNIST dataset for the \textit{recall} modality and from the MNIST dataset for the \textit{transfer learning} modality, in a single-threaded scenario. The models $\varphi_\Theta(\cdot)$ have been encoded with the same $\Theta$ used for the analysis of the accuracy described above. 

The experimental results about the computational time are shown in Table~\ref{tab:timing} where 
\begin{itemize}
    \item $t_c$ is the time spent on the client to generate the keys couple $(k_p, k_s)$, to execute the encryption function $E\left(I, \Theta, k_p\right)$ and the decryption function $D(\hat{y}, \Theta, k_s)$. The machine used for the client is equipped with a 2.30GHz 64-bit dual-core processor and 8192 MB of RAM. 
    \item $t_s$ is the time spent by the server to encode the model $\varphi(\cdot)$ and process the encrypted image, $\varphi_\Theta(\hat{I})$. The machine used as a server is an Amazon EC2 instance with 72 64-bit cores at 3.6GHz and 144 GIB of RAM.
    \item $t_t$ estimates the transmission times of sending the encrypted image $\hat{I}$ and receiving back the encrypted result $\hat{y}$. For the transmission part we modeled an high-bandwidth scenario, where we employ the transmission technology \textit{Wi-Fi 4} (standard IEEE 802.11n) using a single-antenna with 64-QAM modulation on the 20 MHz channel with he data-rate $\rho=72.2Mb/s$~\cite{xiao2005ieee}.
\end{itemize}
Two main comments arise. First, as expected, all the three component of the computational times increase with $m$. More specifically, $t_c$ and $t_s$ increase due to the larger computational load required to process encrypted data with larger $m$, while $t_t$ increases due to the increase of the size of the ciphertexts. In addition, $t_c$ is always lower than $t_s$ since $E\left(I, \Theta, k_p\right)$ and  $D(\hat{y}, \Theta, k_s)$ are  less computational demanding than $\varphi_\Theta(\hat{I})$.
Second, an increase in $p$ does not result in a variation of the computational times $t$s. 
All in all, $p$ should be tuned focusing on the accuracy of the results, while $m$ must be tuned by trading-off accuracy and computational load.
\section{Conclusions}
The aim of this paper was to introduce a novel privacy-preserving distributed architecture for deep-learning-as-service. The proposed architecture, which relies on Homomorphic Encryption, supports the Cloud-based processing of encrypted data to preserve the privacy of user data. The proposed architecture has been tailored to Convolutional Neural Networks and an implementation based on Python and Amazon AWS is made available. Experimental results show the effectiveness of what proposed.

Future work will consider the automatic configuration of the Homomorphic Encryption parameters, the extension of the deep learning models to deep recurrent neural networks and optimized client implementation for Internet-of-Things devices (characterized by constraints on computation and memory).
\label{sct:conclusions}
%

\section*{Acknowledgement}
This work has been partially supported by the project ``GAUChO " Project funded by MIUR under PRIN 2015.



%

\bibliographystyle{IEEEtran}
\bibliography{main}

\end{document}